\documentclass{svjour3}                     
\smartqed  
\usepackage{times}
\usepackage{url}
\usepackage{latexsym}
\usepackage{amsmath}
\usepackage{bm}
\usepackage{tipa}
\usepackage{stfloats}
\usepackage{enumitem}
\usepackage{color}
\usepackage{comment}
\usepackage{graphicx}
\usepackage[small]{caption}
\usepackage{enumitem}
\usepackage{multirow}
\usepackage{booktabs}
\usepackage{wrapfig}
\usepackage{enumitem}
\usepackage[numbers]{natbib}
\usepackage{algorithm}  
\usepackage{algpseudocode}   
\usepackage{amssymb}
\usepackage{amsfonts}
\usepackage{array}
\usepackage{ulem}
\usepackage{rotating}
\usepackage{lscape}
\usepackage{caption}
\usepackage{subfigure}

\usepackage{pgfplots}               
\pgfplotsset{width=9cm,compat=1.13}

\newcolumntype{L}[1]{>{\vspace{0.5em}\begin{minipage}{#1}\raggedright\let\newline\\
\arraybackslash\hspace{0pt}}m{#1}<{\end{minipage}\vspace{0.5em}}}
\newcolumntype{R}[1]{>{\vspace{0.5em}\begin{minipage}{#1}\raggedleft\let\newline\\
\arraybackslash\hspace{0pt}}m{#1}<{\end{minipage}\vspace{0.5em}}}
\newcolumntype{C}[1]{>{\vspace{0.5em}\begin{minipage}{#1}\centering\let\newline\\
\arraybackslash\hspace{0pt}}m{#1}<{\end{minipage}\vspace{0.5em}}}

\newtheorem{mydef}{Problem}

\begin{document}
	
	\fussy
	\sloppy
	\title{\large{A General Framework for Learning Prosodic-Enhanced Representation of Rap Lyrics}}
	 \titlerunning{ }  
	%
	\author{Hongru Liang 
    \and Haozheng Wang 
    \and Qian Li
    \and Jun Wang 
    \and Guandong Xu
    \and Jiawei Chen
    \and Jin-Mao Wei
    \and Zhenglu Yang
    }
	 \authorrunning{} 
	%
	%
	\institute{Hongru Liang \at
                College of Computer Science, Nankai University, China
                \and Haozheng Wang \at
                College of Computer Science, Nankai University, China
                \and Qian Li \at
                College of Computer Science, Nankai University, China
                \and Jun Wang \at
                College of Mathematics and Statistics Science, Ludong University, China
                \and Guandong Xu \at
                Advanced Analytics Institute, University of Technology Sydney, Australia
                \and Jiawei Chen \at
                College of Computer Science, Nankai University, China
                \and Jin-Mao Wei \at
                College of Computer Science, Nankai University, China
                \and Zhenglu Yang* \at
    			College of Computer Science, Nankai University, China\\
                \email{yangzl@nankai.edu.cn}, Tel:~(+86)-22-85388837
                \\
                \\
                *Corresponding author
                }
	\date{}
	\maketitle              
	\vspace*{-3em}
	\begin{abstract}
	Learning and analyzing rap lyrics is a significant basis for many web applications, such as music recommendation, automatic music categorization, and music information retrieval, due to the abundant source of digital music in the World Wide Web. Although numerous studies have explored the topic, knowledge in this field is far from satisfactory, because critical issues, such as prosodic information and its effective representation, as well as appropriate integration of various features, are usually ignored. In this paper, we propose a \underline{h}ierarchical \underline{a}ttention \underline{v}ariational \underline{a}uto\underline{e}ncoder framework~(HAVAE), which simultaneously consider semantic and prosodic features for rap lyrics representation learning. Specifically, the representation of the prosodic features is encoded by phonetic transcriptions with a novel and effective strategy~(i.e., rhyme2vec). Moreover, a feature aggregation strategy is proposed to appropriately integrate various features and generate prosodic-enhanced representation. A comprehensive empirical evaluation  demonstrates that the proposed framework outperforms the state-of-the-art approaches under various metrics in different rap lyrics learning tasks.
	\end{abstract}
	%
	\renewcommand{\algorithmicrequire}{\textbf{Input:}}
\renewcommand{\algorithmicensure}{\textbf{Output:}}
\section{INTRODUCTION}
The explosively increasing availability of digital music in the World Wide Web has inspired broad interest in music learning and analysis~\cite{Malmi2016dopelearning,hu2017medj}. As one of the most popular types of music genres, rap music is worth exploring statistically~\cite{mauch2015evolution}, as it is a basis of many web applications, such as music generation~\cite{Potash2015GhostWriter,wu2013learning}, music information retrieval~\cite{Malmi2016dopelearning}, and scheme identification~\cite{Addanki2013Unsupervised,hirjee2009automatic}. A complicated issue for the task of learning rap lyrics is that rap lyrics are unstructured, thereby hampering the direct use of the off-the-shelf natural language processing techniques in phonological analysis~\cite{wu2013learning}.\par

Recent studies have devoted much attention to rap lyrics by plain text analysis~\cite{Potash2015GhostWriter}, rhyme scheme detection~\cite{Addanki2013Unsupervised,hirjee2009automatic}, rap lyrics generation~\cite{Potash2015GhostWriter,wu2013learning}, and evaluation methodology development~\cite{Hirjee2010Using,Malmi2016dopelearning,potash2016evaluating}. However, the strategies of manipulating features in existing studies are flawed, which dramatically deteriorate the performance of learning rap lyrics. Either partial features~(such as the semantic features obtained from raw lyrics) rather than the complete ones are captured, or ineffective representations of features~(e.g., statistical representations) are employed. The comprehensive and general representations of rap lyrics involving both semantic and prosodic information are urgently required.\par

We notice that variational autoencoder~(VAE)~\cite{kingma2013auto} empirically demonstrates strong power for density modeling and generation studies~\cite{hou2017deep}, including its successful applications in music information retrieval area~\cite{alexey2017music,hadjeres2017glsr}. Nonetheless, these studies are interested in small pitches of music melodies and thereby simply utilize VAEs to reduce the dimensionality of acoustic features. \par

To tackle the aforementioned issues, we scrutinize both semantic and prosodic features in a unified framework specific to rap lyrics learning, that is, a prosodic-enhanced representation of rap lyrics is well constructed in semantic view. Concretely, the semantic information is encoded into vectors through the popular paragraph embedding technology~(i.e., doc2vec)~\cite{quoc2014distributed}. The prosodic features are represented through our newly proposed strategy, dubbed rhyme2vec, in an effective fashion of incorporating various rhyme schemes. Consequently, the prosodic-enhanced representation of rap lyrics is accomplished, in which the VAE-based feature aggregation approach is designed to seamlessly combine the semantic and prosodic information, as well as the attention mechanism is employed to appropriately balance their importance. All of these strategies are integrated into a general representation learning framework named as hierarchical attention VAE network~(HAVAE). The main contributions of this study are as follows:			
\begin{itemize}
	\item We propose a hierarchical attention VAE-based framework, called HAVAE, to effectively address the issue of representation learning for rap lyrics. The framework can represent rap lyrics both prosodically and semantically.
	\item A novel strategy, called rhyme2vec, is proposed to accomplish the prosodic representation learning. This method involves two models, namely, continuous lines and skip-line, to appropriately cope with rhyme schemes with distinct characteristics. 
	\item To effectively integrate various features, we deliberately design a feature aggregation module via VAE network and introduce the hierarchical attention mechanism for seamless fusion of diverse types of information.  
	\item We extensively evaluate the proposed framework on benchmark datasets and fulfill three tasks. Results demonstrate that our framework is remarkably better than the state-of-the-art approaches under various evaluation metrics.
\end{itemize}\par
\section{RELATED WORK}
The most relevant task to learn rap lyrics representation is poetry analysis. Poetry is a highly varied genre, and each poetry type has its distinct structural, rhythmical, and tonal patterns. Rule-based and template-based approaches~\cite{oliveira2012poetryme} were mainstays in the early days. Subsequently, generic algorithm~\cite{ruli2012using}, summarization framework~\cite{rui2013i}, and statistical machine translation models~\cite{he2012generating} were developed. Recent work illustrated that poetry analysis eliminated man-made constraints and professional silo restrictions with the aid of neural networks~\cite{wang2016chinese-song}.\par
Rap lyrics analysis is generally more challenging than poetry analysis, because its structure is more free-form. Hirjee and Brown~\cite{hirjee2009automatic} proposed a probabilistic scoring model based on phoneme frequencies in rap lyrics. Although their model automatically identifies internal and line-final rhymes, it requires additional manually annotated rap lyrics and rhyming pairs. Wu \textit{et al}.~\cite{wu2013learning} explored a generation task in rap battle improvisation. They presented improvisation as a quasi-translation task, in which any given challenge was ``translated'' into a response. However, a ``translation lexicon'' is required before training and the correlation relationship between challenge and response lines are strictly limited to one-to-one correspondence. As such, it is nearly inapplicable in practice. Potash \textit{et al}.~\cite{Potash2015GhostWriter} applied long short-term memory network in generating lyrics that were similar in style to that of a certain rapper, and they presented computational and quantitative evaluation methods. An inevitable drawback is that this system requires sufficient training data to capture the rhythmic style, and thus, rhyming pairs must appear frequently enough in the corpus.\par

DopeLearning, which was proposed by Malmi \textit{et al}. ~\cite{Malmi2016dopelearning}, has been proven to be able to generate powerful features. The authors introduced three kinds of prosodic features, a structural feature, and four kinds of semantic features. DopeLearning has shown desirable performance in rap lyrics learning due to its comprehensive consideration of various features, but it still suffers from a major issue that hampers its effective utilization in practice, that is, a unified representation of rap lyrics has not been generated. Specifically, different types of features are separately extracted to measure different similarity/distance scores for specific tasks. A most crucial issue for rap lyrics representation, i.e., the non-linear relationship among features, is neglected. Furthermore, three prosodic features are extracted simply in statistical views by merely considering vowel phonemes, and apparently, these features cannot represent the entire prosodic information in rap lyrics.\par

As for feature aggregation, VAE has been discussed extensively in computer vision~\cite{Dosovitskiy2016Generating} and nature language processing~\cite{chen2017kate}. For example, in music information retrieval area, several VAE-based frameworks have been proposed to solve specific problems. Fabius and Amersfoort~\cite{fabius2014variational} proposed a variational recurrent autoencoder (VRAE) to generate video game melodies. The work of Alexey and Ivan~\cite{alexey2017music} explored a history supported VRAE to generate monotonic music. Hadjeres \textit{et al}.~\cite{hadjeres2017glsr} introduced a refined regularization function for VAE to generate chorales polyphony. The aforementioned works focus on music melodies bypassing music lyrics, and correspondingly, VAE is only deployed to model plain audio features. Furthermore, the VAE method is constrained to be a generative tool in the conventional sense.\par

As such, we propose an effective VAE-based framework that embeds both semantic and prosodic information. Intuitively, we expect to extract the representative prosodic information and model the relationships between the semantic and prosodic features. The prosodic features are generated by an effective strategy, namely, rhyme2vec. Specifically, we use the VAE network to integrate both prosodic features and semantic features, which are generated via rhyme2vec and doc2vec, respectively. The final representation of rap lyrics is prosodic-aware in addition to semantic awareness. 

\begin{figure*}[t]
	\centering 
	\includegraphics[width=8.88cm]{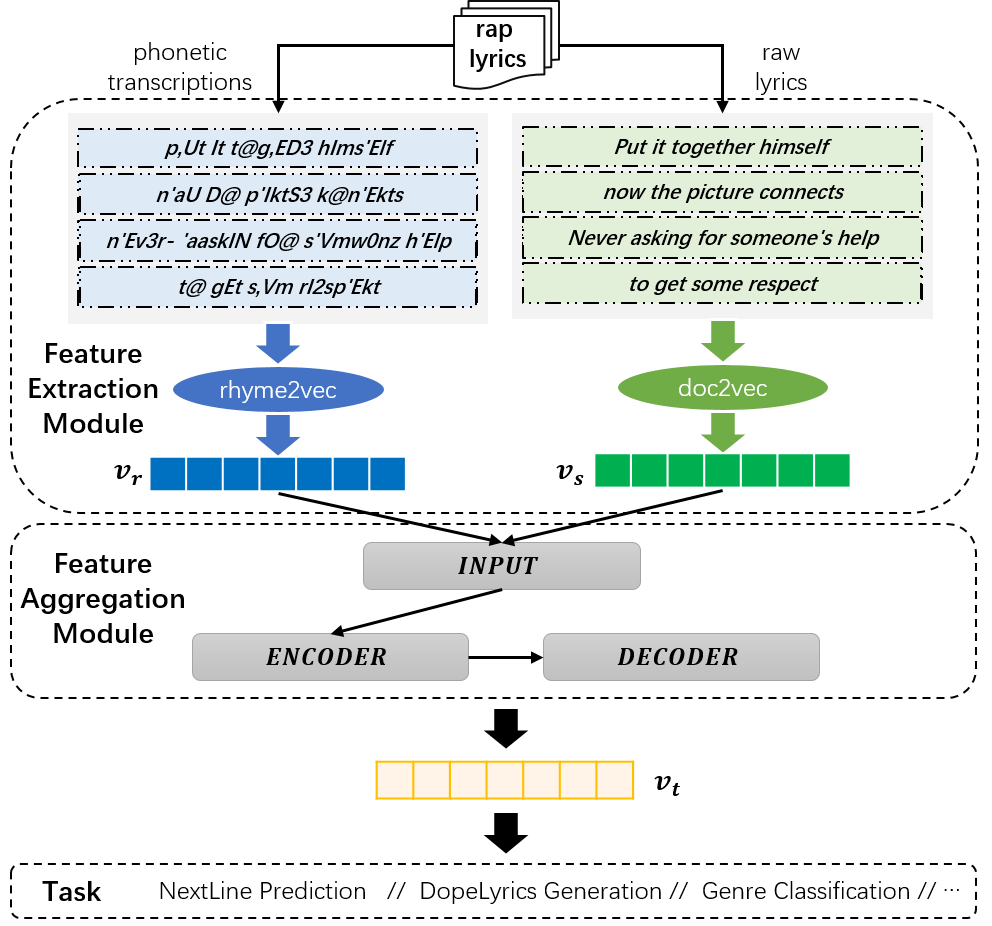}
	\caption{Architecture of HAVAE}
	\label{fig:model}
\end{figure*}

\section{MODEL DESCRIPTION}
The architecture of the proposed framework is illustrated in Fig.~\ref{fig:model}. It consists of two main modules, namely, feature extraction and feature aggregation. The feature extraction module encodes the input materials, which involve the phonetic transcriptions and the raw lyrics, into distributed prosodic and semantic vectors. The feature aggregation module fuses both information and learns the prosodic-enhanced representation vectors. The learned vectors can be used to represent the rap lyrics in various tasks, such as NextLine prediction. The entire structure is a hierarchical attention variational network, and we name it as HAVAE.

\subsection{Preliminary}
\label{section:Preliminary}
Table~\ref{tab:notation} presents an overview of useful notations in this paper. Vectors, matrices, and sets are highlighted in bold. Assume that $\bm S$ indicates a rap song, which consists of $k$ rap lines. Let $\bm {L}=\{l_{i}\}_{i=1}^{k}$ denote the raw lyrics of $\bm S$, and $\bm {{P}}=\left\{p_{i}\right\}_{i=1}^{k}$ denote the corresponding phonetic transcriptions of $\bm S$, respectively. For ease of exposition, we present a concrete example in Table~\ref{tab:example}~(i.e., four consecutive rap lines from Fort Minor's \textit{Remember the Name}) to illustrate the intuitive idea and explain the proposed techniques. We use $\bm {v_s}$ to represent the semantic vector encoded from $\bm {L}$, and $\bm {v_r}$ to represent the prosodic vector encoded from $\bm {P}$. $\bm {v_t}$ denotes the target unified representation of $\bm S$ learned by the proposed feature aggregation method. The number of samples is represented as $N$. \par

\begin{table}[t]
\setlength{\abovetopsep}{0.5ex}
	\setlength{\belowrulesep}{0pt}
	\setlength{\aboverulesep}{0pt}
	\centering
	\caption{Overview of useful notations}
	\scalebox{1.0}{
		\begin{tabular}{|l|p{0.8\textwidth}|}
			\toprule
			\textbf{Symbols} & \textbf{Descriptions}\\
			\midrule\midrule
			$\bm S,\bm{L}, \bm{P}$ &  a rap song, and its raw lyrics and corresponding phonetic transcriptions\\
			\midrule            
			$k$& number of lines in $\bm S$\\
			\midrule
			${l_i},{p_i}$& $i$-${th}$ line of $\bm{L}$ and $\bm{P}$\\
			\midrule
			$\bm{v_s},\bm{v_r},\bm{v_t}$&semantic vector of $\bm{L}$ , prosodic vector of $\bm{P}$ , and target representation of $\bm S$  \\
			\midrule
			$\bm{B}$&prosodic block\\
			\midrule
			$\bm{\rho}$& prosodic vector of $\bm{B}$\\
			\midrule
			$\bm{\psi}$& dense vector of a phoneme\\
			\midrule
			$\bm{\xi}$& scheme vector of $\bm{B}$\\
			\midrule
			$n$& number of phonemes in a prosodic block\\
			\midrule
			$w$& size of the sliding window over prosodic blocks\\
			\midrule
			$N$& number of samples\\
			\bottomrule
		\end{tabular}%
	}
	\label{tab:notation}%
\end{table}%
\begin{table}[t]	
	\centering
	\caption{Rap lyrics with their phonetic transcriptions}
	\begin{tabular}{p{0.5\textwidth}l}
		\toprule
		\textbf{raw lyrics ($\bm{L}$)} & \textbf{phonetic transcription ($\bm{P}$)}\\
		\midrule
		$[l_1]$~Put it together himself & $[p_1]$~\color{red}{p,Ut It t@g,ED3 hIms'Elf} \\
		$[l_2]$~Now the picture connects & $[p_2]$~\color{blue}{n'aU D@ p'IktS3 k@n'Ekts} \\
		$[l_3]$~Never asking for someone's help & $[p_3]$~\color{red}{n'Ev3r- 'aaskIN fO@ s'Vmw0nz h'Elp} \\
		$[l_4]$~to get some respect & $[p_4]$~\color{blue}{t@ gEt s,Vm rI2sp'Ekt} \\
		\bottomrule
	\end{tabular}%
	\label{tab:example}%
\end{table}%

The English language is known to have 48 International Phonetic Alphabets~\cite{international1999handbook}. We treat affricates~(e.g., [\textteshlig]) and diphthongs~(e.g., [\textopeno\textsci]) as two separate phonemes, and translate the rap lyrics into phoneme codes using the eSpeak tool~\cite{duddington2012espeak}. The final phonetic transcription file is a sequence of characters from a token-level alphabet and each token acts as an individual member. The alphabet is defined as follows:
\begin{equation*}
\bm{pbtTdDsSzZkgfvhmnNljwr2Y3L5aAeEiI0VuUoO}.
\end{equation*}\par
Based on the phonetic transcription files with phoneme codes, we can perform examinations w.r.t. an important factor in rap lyrics learning, that is, rhyme. A rhyme is a repetition of similar sounds~(or the same sound) in two or more words~\cite{bryant1990rhyme} and may appear in a single rap line or cross lines, and it can be classified into two mainstream rhyme schemes, namely, monorhyme\footnote{ Monorhyme is a rhyme scheme in which each line has an identical rhyme.} and alternate rhyme\footnote{In alternate rhyme, the rhyme is on alternate lines.}. Monorhyme, alternate rhyme, and other schemes are randomly throughout an entire rap song, exhibiting different levels of importance. \par

We assume that the rap lines have both monorhyme and alternate rhymes. For monorhyme, we unify all consecutive lines as one prosodic block~($\bm{B_m}$). Regarding alternate rhyme, we split rap lines into two prosodic blocks. One block includes all of the odd lines~($\bm{B_o}$, i.e., the red lines in Table~\ref{tab:example}), and the other block includes all of the even lines~($\bm{B_e}$, i.e., the blue lines in Table~\ref{tab:example}). The number of prosodic blocks is triple of the number of samples~(i.e., $3N$). We define the problem of learning prosodic representation of rap lyrics as follows:
\begin{mydef}
	\label{p:rhyme}
	Given a sequence of $k$ rap lines $\bm S$, $\bm {P}$ is its phonetic transcription. Let $\bm {\rho_{m}}$ denote the monorhyme representation vector generated from $\bm {B_{m}}$, and $\bm {\rho_{a}}$ denote the alternate rhyme representation vector generated from $\bm {B_{o}}$ and $\bm {B_{e}}$. The goal of learning the prosodic representation of rap lyrics is to generate the prosodic vector $\bm {v_r}$ of $\bm {P}$ by combining $\bm {\rho_{m}}$ and $\bm {\rho_{a}}$.
\end{mydef}
\par
To solve Problem~\ref{p:rhyme}, we propose a novel and robust model in Section~\ref{sec:fe}, called rhyme2vec. This model involves two sub-models: the continuous lines model handles the monorhyme w.r.t. $\bm {B_{m}}$, and the skip-line model handles the alternate rhymes w.r.t. $\bm {B_{o}}$ and $\bm {B_{e}}$.  
\par
On the basis of analyzing the prosodic representation of rap lyrics, we further discuss the task of learning its complete representation. A good rap song possesses an outstanding topic and maintains catchy rhymes. Thus, an appropriate representation of rap lyrics needs to involve both semantic and prosodic information. Here, we provide the definition of the task for rap lyrics representation learning as follows:
\begin{mydef}
	\label{p:all}
	Given a rap song $\bm S$ with a sequence of $k$ rap lines, and assuming that its prosodic vector $\bm {v_r}$ and semantic vector $\bm {v_s}$ are known, the goal of learning the representation of rap lyrics is to generate $\bm v_t$ by combining the useful information of $\bm {v_r}$ and $\bm {v_s}$.
\end{mydef}
\par
An attentive VAE-based feature aggregation module is proposed in Section~\ref{utm} to combine $\bm {v_s}$ and $\bm {v_r}$ seamlessly. In addition, the attention mechanism is introduced to model the mutual relationship between $\bm {v_s}$ and $\bm {v_r}$. Consequently, the target representation $\bm {v_t}$ is learned by sampling on a latent Gaussian distribution, and it is expected to be utilized in various rap lyrics-based tasks, as demonstrated in Section~\ref{exp}.\par

\subsection{Feature Extraction Module}
\label{sec:fe}
The feature extraction module (illustrated as the first dotted box in Fig.~\ref{fig:model}) is utilized to generate fine-grained features . This module consists of prosodic and semantic sections.\\

\begin{figure}[t]
	\centering
	\includegraphics[width=0.8\textwidth]{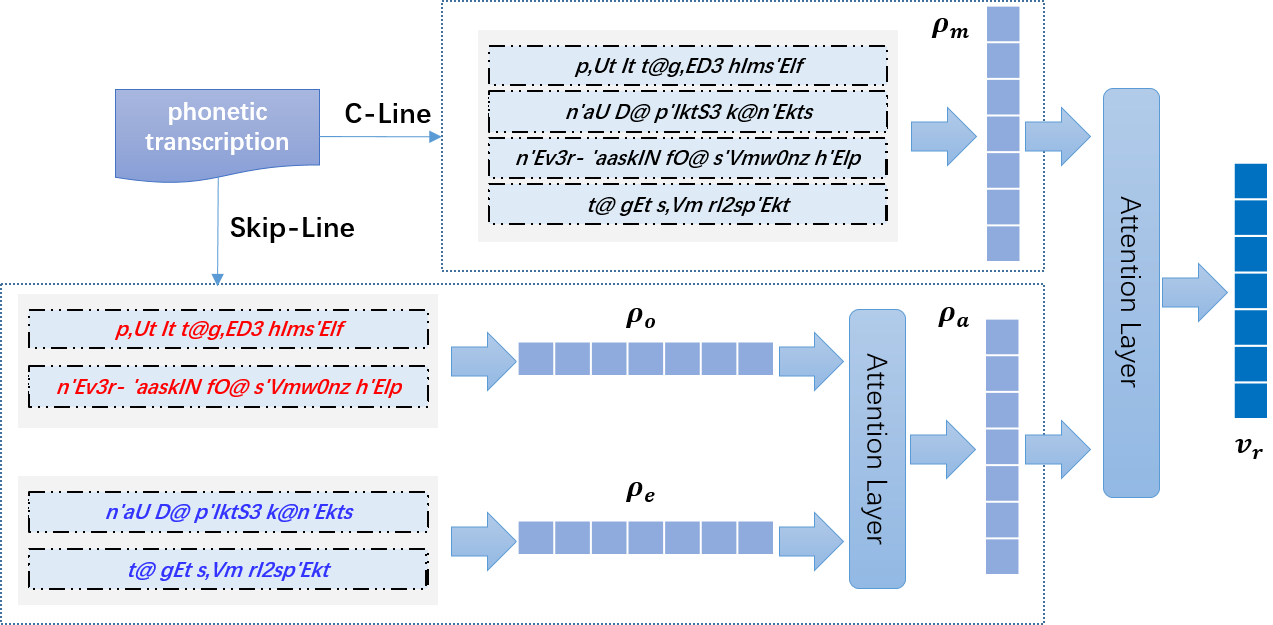}
	\caption{Architecture of rhyme2vec}
	\label{fig:rhyme2vec}
\end{figure}

\noindent{\bf Prosodic Section.}
In the prosodic section, we introduce a novel method to tackle Problem~\ref{p:rhyme}. Fig.~\ref{fig:rhyme2vec} illustrates the full process to obtain the prosodic representation $\bm {L}$ of given rap lyrics. We call our rhyme embedding method rhyme2vec.\par
Inspired by the idea of modeling document~\cite{lei2017swim,lei2018sequicity,quoc2014distributed}, for each prosodic block, we import a scheme vector, which is denoted as $\bm {\xi}$. As aforementioned, each rap song possesses three prosodic blocks. We first map each phoneme into a unique vector, denoted as $\bm {\psi _i}$. All phonemes in the same prosodic block share the same $\bm \xi$. The embedding of a prosodic block is denoted as $\bm {\rho}$. $\bm {\rho}$ is defined as the sum of the phoneme vectors and the scheme vector, formulated as $\bm {\rho}=\sum_{i=1}^{n}\bm {\psi_{i}}+\bm{\xi}$. Given a piece of $\bm P$ with a set of phoneme vectors $\bm \psi$ and scheme vectors $\bm \xi$, we maximize its log probability. The objective function is written as
\begin{equation}
\label{eq:pho}
\max\limits_{\bm {\psi},\bm {\xi}}\mathcal {F}_{\bm r} =\sum\limits_{i=w}^{n-w}\frac{1}{{n}-2w}\log {p}\left(\bm {\psi_{i}}|\bm {\psi_{i-w}}: \bm {\psi_{i+w}}, \bm {\xi}\right).
\end{equation}
\par
Inspired by \cite{lei2018linguistic,lei2019revisit}, we deploy the negative sampling strategy\cite{mikolov2013efficient} for effective training. Let $\bm{P_0}$ denote the negative sample set, and $\bm{P_1}$ denote the positive sample set, which is the same as the original set. Let $\bm {c_i}$ indicate the context content~($\left\{\bm {\psi_{i-w}}: \bm {\psi_{i+w}}, \bm {\xi}\right\}$) of $\bm{\psi_i}$, for each positive sample $\bm{p}=\left[\bm{\psi_i}|_{i=w,...,n-w},\bm {c_i}\right] \in \bm{P_1}$,
we randomly construct $\zeta$ negative samples $\{\bm{c}|{[\bm{\psi_i},\bm{c}]\notin\bm{P_1}}\}$. $\bm{P_0}$ is the set of negative samples and is $\zeta$ times larger than $\bm{P_1}$, with holding $\bm{P_0} \cap \bm{P_1}=\bm {\varnothing}$. We use $\bm{P_1}\cup\bm{P_0}$ as the entire corpus, and the objective function can be rewritten as
\begin{equation}
\label{eq:rhyme}
\max\limits_{\bm {\psi},\bm {\xi}}{\left[{\mathcal{F}_{r}}(\bm{P_1})- {\mathcal{F}_{r}}(\bm{P_0})\right]}.
\end{equation}
The first term refers to the log probability of the positive set~($\bm{P_1}$), and the second term refers to the log probability of the negative set~($\bm{P_0}$).\par
\begin{itemize}
\item \noindent{\textit{Algorithm Analysis.} }
We summarize the embedding procedure of prosodic blocks in Algorithm~\ref{alg:prosodic}. In line~\ref{l:init}, we label every prosodic block with its index, which serves as a pseudo phoneme of rhyme scheme; then, we expand the training data set with negative samples in lines [\ref{2:negative}-\ref{7:negative}]; after that, for each sample in the combined corpus with both positive and negative ones, we obtain its phoneme vectors and scheme vector by optimizing Eq.~(\ref{eq:rhyme}) in lines [\ref{8:optimization}-\ref{13:optimization}]; finally, the representation of the prosodic blocks~(i.e., $\{\bm{\rho}\}$) is obtained in lines [\ref{a1:infer}-\ref{a1:endinfer}], and outputted in line \ref{14:return}.
\par
In terms of time consumption, it takes about $\mathcal{O}(\zeta nN)$ for the negative sampling. The convergence state can be reached by predefining the iteration time $t$, and correspondingly, to obtain the representation vectors $\{\bm{\rho}\}$ in lines [\ref{8:optimization}-\ref{a1:endinfer}] approximately costs $\mathcal{O}(tn^2+tN^2)$. Thus, the total time of Algorithm~\ref{alg:prosodic} is $\mathcal{O}(\zeta nN+tn^2+tN^2)$, where $n$ and $N$ are the numbers of phonemes and samples, respectively. {{Commonly, $N$ is much larger than $n$ and $\zeta$, hence, the time complexity is approximately equal to $\mathcal{O}(\zeta nN+tN^2)$}}.
\begin{algorithm}[t]
	\caption{{Embedding procedure of prosodic blocks}} 
	\label{alg:prosodic}
	\begin{algorithmic}[1]
	\Require{prosodic blocks $\{\bm{B_i}\}_{i=1}^{3N}$, window size $w$, and the number of negative samples $\zeta$}
	\Ensure{representation of the prosodic blocks $\{\bm{\rho}\}$}	
	\State $\bm{P_{1}}=\{{[\bm{B_i}:i]}\}_{i=1}^{3N},\bm{P_{0}}=\varnothing$ \label{l:init};
	{\For {each $\bm{p}\in \bm{P_{1}}$}\label{2:negative}
	\For {each phoneme $\bm {\psi_i}|_{i=w,...,n-w}$ in $\bm{p}$}
	\State randomly construct $\zeta$ negative samples $\{\bm{c}|{[\bm{\psi_i},\bm{c}]\notin\bm{P_1} \wedge [\bm{\psi_i},\bm{c}]\notin\bm{P_0}}\}$;
	\State $\bm{P_0}=\bm{P_0} \bigcup \{{[\bm{\psi_i},\bm{c}]}\}$;
	\EndFor
	\EndFor\label{7:negative}
\While{not converged}\label{8:optimization}
	    \For {each $\bm{p}\in \bm{P_{0}} \bigcup \bm{P_{1}}$}    	
			\State compute $\{\bm {\psi_i}\}_{i=1}^{n}$ and $\bm{\xi}$ by optimizing Eq.~(\ref{eq:rhyme}); 
    	\EndFor
\EndWhile\label{13:optimization}
\For {each $\bm{p}\in \bm{P_1}$}\label{a1:infer}
    		 \State $\bm{\rho}=\sum_i\bm{\psi_i}+\bm{\xi}$;
    	\EndFor\label{a1:endinfer}

\State\Return $\{\bm{\rho}\}$;}\label{14:return}
	\end{algorithmic}	
\end{algorithm}
\end{itemize}

\par
Specifically, to describe the monorhyme, we propose the continuous lines model, which is denoted as C-Line for short. In C-Line (illustrated in the upper part of Fig.~\ref{fig:rhyme2vec}), we regard the input consecutive lines~($\bm{B_m}$) as a content, and generate a distributed vector $\bm{\rho_m}$ from the corresponding phoneme vectors and prosodic block vector using the abovementioned method. \par

As to the alternate rhyme, we use $\bm {\rho_a}$ to indicate the dense representation. Seeing that monorhyme and alternative rhyme have different degrees of importance in rap lyrics, we introduce the Attention Layer, which assigns appropriate weights for $\bm {\rho_m}$ and $\bm {\rho_a}$, respectively. The Attention Layer is defined as $\bm{AttL}:\bm{\mathcal{H}:=[\bm{h_i}]_{i=1}^{m}\in\mathbb{R}^{d\times m}\to \bm{t}\in\mathbb{R}^{d}}$, detailedly expressed as follows:
\begin{equation}
\bm{t}= \sum_{i=1}^{m}{\alpha{_i}\bm{ h{_i}}},\\
\alpha_i= \frac{exp\left( \bm{\hat{h}{_i}^{\top } \hat{h}_{c}}\right) }{\sum_{j=1}^{m}{exp\left( \bm {\hat{h}{_j}^\top \hat{h}_{c}} \right) } },\\
\bm{\hat{h}_i}= tanh(\bm {W}\bm{h_i}+\bm{b}),
\end{equation}
where $\bm{\mathcal{H}}$ is the input matrix and built by a stack of $m$ vectors $\bm{h_i}$, $\bm t$ is the target vector learned by $\bm{AttL}$, $\alpha_i$ is the attention weight scalar of $\bm{h_i}$ and evaluated by the latent vector $\bm{\hat{h}_i}$ of $\bm{h_i}$ along with the randomly initialized context vector $\bm{\hat{h}_c}$, and $W$ and $b$ are hyper-parameters involved in the $tanh$ function to compute $\bm{\hat{h}_i}$. Hence, we formulate the prosodic vector as $\bm{v_r}=\bm{AttL}~(\bm{\mathcal{H}:=[\bm{\rho_{m}};\bm{\rho_a}]})$.\par

Furthermore, to describe the alternative rhyme comprehensively, we propose the skip-line models, which is denoted as Skip-Line for short. Unlike in C-Line, we first split the phonetic transcription into the odd~(with all the odd lines, i.e., $\bm{B_o}$) and even prosodic blocks~(with all the even lines, i.e., $\bm{B_e}$) in Skip-Line (illustrated as the lower part in Fig.~\ref{fig:rhyme2vec}). $\bm{\rho_o}$ and $ \bm{\rho_e}$ are encoded from $\bm{B_o}$ and $\bm{B_e}$ as $\bm{\rho_m}$. In addition, similar to aggregating monorhyme and alternate rhyme vectors, we employ the Attention Layer on $\bm{\rho_o}$ and $\bm{\rho_e}$, that is, $\bm{\rho_a}=\bm{AttL}(\bm{\mathcal{H}}:=[\bm{\rho_o};\bm{\rho_e}])$.\par

The final prosodic representation of $\bm {P}$ is obtained through the proposed rhyme2vec method and reformulated as $\bm{v_r}=\bm{AttL}~(\bm{\mathcal{H}}:=[\bm {\rho_m}; \bm{AttL}~(\bm{\mathcal{H}}:=[\bm{\rho_o}; \bm{\rho_e}])])$.\\

\noindent{\bf Semantic Section.}
To extract semantic features, we employ the doc2vec approach~\cite{quoc2014distributed}. Doc2vec is a state-of-the-art sentence embedding technique, which stems from the distributed memory model of paragraph vectors (PV-DM) and the distributed bag-of-words version of paragraph vectors (PV-DBOW). The empirical analysis has shown that PV-DM usually works better than PV-DBOW~\cite{quoc2014distributed}, and thus in this paper, we deploy PV-DM to generate the semantic vector of $\bm {L}$, i.e., $\bm {v_s}$. At line level, we regard consecutive lines as a paragraph. At song level, we regard each rap song as a document.\par 

\subsection{Feature Aggregation Module}
\label{utm}
\begin{figure}[t]
	\centering
	\includegraphics[width=0.7\textwidth]{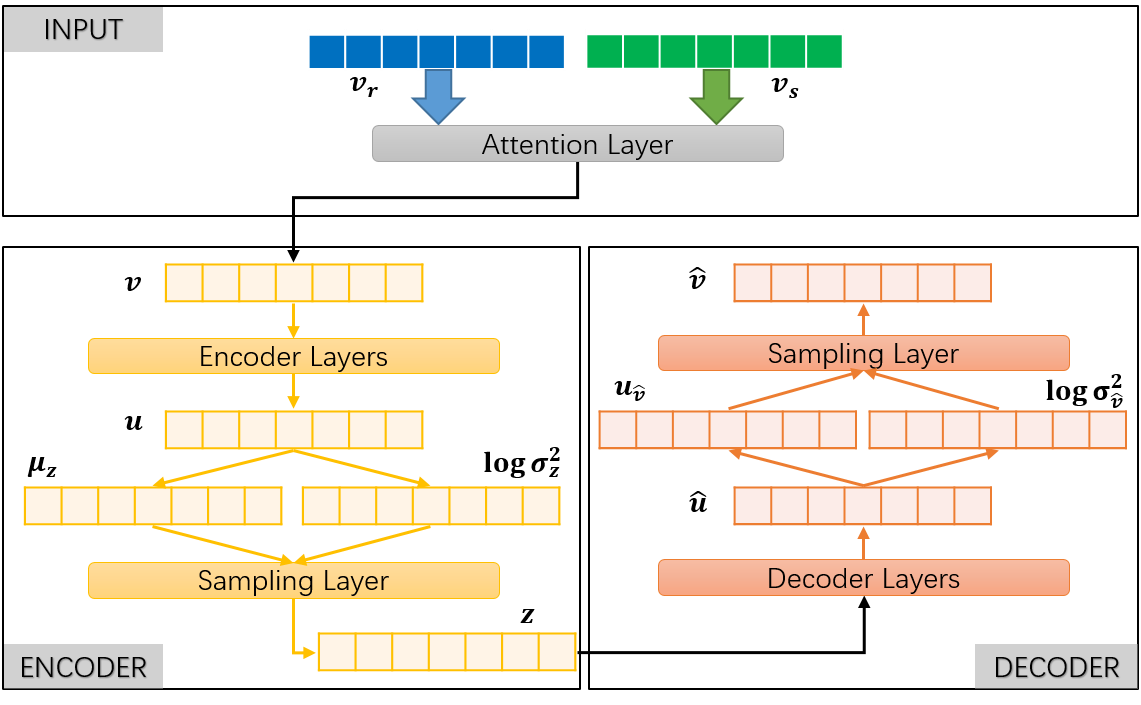}
	\caption{Architecture of the feature aggregation module}
	\label{fig:aggregation}
\end{figure}

The feature aggregation module is designed to solve Problem~\ref{p:all}, and is an essential part of our model~(illustrated as the second dotted box in Fig.~\ref{fig:model}). In this section, we design a VAE based network to produce representation of rap lyrics by combining prosodic and semantic information. The architecture of the proposed VAE based network is similar to the basic autoencoder, while the difference is that the former generates additional noises with regard to a probability distribution. This is based on an intuitive idea that the generated noises can benefit the feature aggregating process. Moreover, we introduce the attention mechanism to balance their importance. Fig.~\ref{fig:aggregation} illustrates the structure, which consists of three main stages, namely, INPUT, ENCODER, and DECODER.\par
In the INPUT stage~(see upper part in Fig.~\ref{fig:aggregation}), we use the aforementioned Attention Layer to handle the input variables. $\bm{v_r}$ and $\bm{v_s}$ are the fine-grained features of prosodic and semantic information, and have already been learned through the feature extraction module. The output vector $\bm v$ of the Attention Layer in this stage is utilized as the input of the following stage. We formulate this procedure as $\bm v=\bm{AttL}(\bm{\mathcal{H}}:=[\bm{v_r};\bm{v_s}])$.\par
In the ENCODER stage, we first deploy the Encoder Layers over the input vector~($\bm v$). The Encoder Layers are a sequence of densely-connected layers, and the $i$-${th}$ densely-connected layer is formulated as $\bm{v_{i}}=\bm{\delta}(W \bm{v_{i-1}}+b)_{i=2}^{\chi}$, where $\chi$ is the number of the layers, $\bm \delta$ is an activation function, and $\bm{v_1}:=\bm v$. The Encoder Layers generate a latent representation of $\bm v$. The VAE network generates latent variable $\bm z$ on the basis of $\bm v$, and $\bm z$ is expected to be most representative for $\bm v$. Suppose $\bm z$ follows the Gaussian distribution ${\mathcal{N}(\bm {\mu_z},\bm{\sigma_z^2})}$, where $\bm{\mu_z}$ indicates the mean vector, and $\log\bm{\sigma_z^2}$ indicates the log-covariance vector~($\bm{\sigma_z}$ indicates the standard deviation vector). For telesis, we use the latent representation, which is denoted as  $\bm u $, to generate the mean vector and the log-covariance vector of $\bm z$ through fully-connected layers. For ease of calculating the gradient, ``reparameterization trick''~\cite{kingma2013auto} is adopted. Mathematically, the Sampling Layer is formulated as
\begin{equation}
\bm{z}=\bm{\mu_z}+\bm{\sigma_z}\odot \bm{\epsilon},\quad \bm \epsilon \sim \bm{\mathcal{N}(0, \bm 1)};
\end{equation}
where $\bm \epsilon$ is the noise variable following the normal distribution, and $\odot$ represents the element-wise product. 
\par
The DECODER stage accomplishes an inverse manipulation of the ENCODER stage. $\bm z$ is utilized as the input of this stage. We then use the Decoder Layers, which are a sequence of densely-connected layers with reversed order of the Encoder Layers, to decode $\bm z$ to $\bm{\hat{u}}$\footnote{The dimension of $\bm{\hat u}$ is equal to that of $\bm u$.}. On the basis of $\bm{\hat{u}}$, we can obtain the mean vector $\bm{\mu_{\hat{v}}}$ and the standard deviation vector $\bm{\sigma_{\hat{v}}}$ of $\bm{\hat{v}}$ with the normal distribution ${\mathcal{N}(\bm {\mu_{\hat v}},\bm{\sigma_{\hat v}^2})}$. $\bm{\hat v}$ can be considered as the reconstructed vector of $\bm v$. 
We use $\bm{\mu_z}$ as the target representation of the rap lyrics, that is, $\bm{\mu_z}=\bm{v_t}$. \\

\noindent{\bf Loss Function.}
Inspired by~\cite{Kingma2014Semi}, we incorporate the label information. We construct a latent variable $\bm {\hat{y}}$ on the basis of $\bm {v_t}$ to represent the true label set $\bm y$. The operation can be expressed as $\bm {\hat{y}}=sigmoid\left(\bm {v_t}\right)$. $\bm {\hat y}$ and $\bm y$  possess the same dimension. \par
Based on the above analysis, the eventual objective function of our feature aggregation module is given as
\begin{equation}
\label{eq:total}
\min\limits_{\bm {v_t},\bm {\hat{y}}}(\mathcal {F}_{vae}+\alpha\mathcal {F}_{label}),
\end{equation} 
where $\mathcal {F}_{vae}$ is the loss of the VAE-based feature aggregation network, $\mathcal {F}_{label}$ is the loss between the latent variable $\bm {\hat{y}}$ and the true label set $\bm y$, and $\alpha$ is a hyper parameter to balance the importance of $\mathcal {F}_{vae}$ and $\mathcal {F}_{label}$.\par

$\mathcal {F}_{vae}$ can be divided into two parts, and formulated as $\mathcal {F}_{vae}=\mathcal{F}_{GL}+\mathcal{F}_{RL}$. The first term describes the generative loss, evaluating the similarity between the latent variable~($\bm z$) and the original data~($\bm{v_r}$ and $\bm{v_s}$). The second term represents the reconstructed loss, which  estimates the loss between the latent variable~($\bm z$) and the reconstruction vector~($\bm{\hat{v}}$).\par

Let $ \mathcal {Q}( \bm z|\bm v)$ denote an approximate posterior distribution of $ \bm z $ and $ \mathcal{P}\left(\bm z\right) $ denote the prior distribution. Intuitively, we expect to train our VAE network by minimizing the difference between~$\mathcal Q (\bm z|\bm{v})$ and $\mathcal P\left(\bm z\right) $. In doing so, we can capture the most representative vector $\bm z$  for $\bm v$ with minimal difference between both variables. We introduce the Kullback-Leibler $(\mathcal{KL})$ divergence metric to evaluate the difference between $\bm z$  and $\bm v$, and mathematically expressed as follows: 
\begin{equation}
\label{eq:kl}
{{\mathcal{KL}}\left[\mathcal{Q}(\bm z|\bm v)||\mathcal{P}(\bm z)\right]=\sum _{i=1}^{N}\mathcal{Q}(\bm {z_i}|\bm {v_i}) \times \log { \left[\frac{\mathcal{Q}(\bm {z_i}|\bm {v_i})}{\mathcal P\left(\bm z_i\right)}\right]}.}
\end{equation}
Suppose both the prior and posterior approximations are Gaussian distributed. The following equation holds:
\begin{equation}
{\mathcal{F}_{GL}=-{\mathcal{KL}}\left[\mathcal{Q}(\bm z|\bm v)||\mathcal{P}(\bm z)\right]=\frac{1}{2}\sum_{i=1}^{N}\left[1+\log(\bm{\sigma_z})_i^2-(\bm{\mu_z})_i^2-(\bm{\sigma_z})_i^2\right].}
\end{equation}\par
The reconstructed loss describes the loss between the latent variable and the reconstruction of the input. Seeing that the decoder is a multivariate Gaussian with a diagonal covariance structure, we introduce the reconstructed loss, which is defined as $\mathcal{F}_{RL}=\log\mathcal{P}(\bm v|\bm z)=log\mathcal{N}(\bm v; \bm{\mu_{\hat{v}}}, \bm{\sigma_{\hat{v}}^2})$. \par
In conclusion, the loss function of the VAE network is formulated as
\begin{equation}
\mathcal{F}_{vae}=-{\mathcal{KL}}[\mathcal{Q}(\bm z|\bm v)||\mathcal{P}(\bm z)]+\log\mathcal{P}(\bm v|\bm z).
\end{equation}
\par
Moreover, $\mathcal{F}_{label}$ can be estimated through the binary cross entropy as follows:
\begin{equation}
\mathcal{F}_{label}=-\frac{1}{NM}{\sum_{i=1}^N}{\sum_{j=1}^{M}}[y_{ij}{\log \hat{y}_{ij} }+(1-y_{ij})\log(1-\hat{y}_{ij})],
\end{equation}
where $M$ is the dimension of the label set.\\

\begin{algorithm}[t]
	\caption{{Feature aggregation procedure of HAVAE}} 
	\label{alg:fa}
	\begin{algorithmic}[1]
		\Require{prosodic vector $\bm{v_r}$, semantic vector $\bm{v_s}$}, and true label set $\bm y$
		\Ensure{the hyper-parameter settings for generating unified representation}	
        \State {randomly initialize $\{\bm{W_i^{E}}\}_{i=1}^{\chi{+1}}, \{\bm{b_i^{E}}\}_{i=1}^{\chi{+1}}, \{\bm{W_i^{D}}\}_{i=1}^{\chi{+1}}, \{\bm{b_i^{D}}\}_{i=1}^{\chi{+1}}, \bm{W}^{y}, \bm{b}^{y}$;\label{a2:initialization}
		\While{not converged} \label{a2:whileB}
        \State /*INPUT*/
		\State $\bm{v}=\bm{AttL(\bm{\mathcal{H}:={[\bm{v_r},\bm{v_s}]}})}$;\label{a2:input}        
		\State /*ENCODER*/
		\State $\bm{v_1}:=\bm v$; \label{a2:encodeB}
		\For {$i=2$ to $\chi$}
		\State $\bm{v_{i}}=\bm\delta(\bm{W_{i-1}^{E}}\bm{v_{i-1}}+\bm{b_{i-1}^{E}})$;
		\EndFor
		\State $\bm{\mu_z}=\bm\delta(\bm{W_{\chi}^{E}}\bm{v_\chi}+\bm{b_{\chi}^{E}})$,~$\log\bm{{\sigma_z^2}}=\bm\delta(\bm{W_{\chi{+1}}^{E}}\bm{v_\chi}+\bm{b_{\chi{+1}}^{E}})$;\label{a2:target}
		\State randomly construct $\bm \epsilon \sim \bm{\mathcal{N}(0, \bm 1)}$;
		\State $\bm z=\bm{\mu_z}+\bm{\sigma_z}\odot \bm{\epsilon}$;\label{a2:encodeE}
		\State /*DECODER*/ 
		\State $\bm{z_1}:=\bm z$;\label{a2:decodeB}
		\For {$i=2$ to $\chi$}
		\State $\bm{z_{i}}=\bm\delta(\bm{W_{i-1}^{D}}\bm{z_{i-1}}+\bm{b_{i-1}^{D}})$;
		\EndFor
		\State $\bm{\mu_{\hat{v}}}=\bm\delta(\bm{W_{\chi}^{D}}\bm{z_\chi}+\bm{b_{\chi}^{D}})$,~$\log\bm{{\sigma_{\hat{v}}^2}}=\bm\delta(\bm{W_{\chi{+1}}^{D}}\bm{z_\chi}+\bm{b_{\chi{+1}}^{D}})$;
		\State randomly construct $\bm{\hat \epsilon} \sim \bm{\mathcal{N}(0, \bm 1)}$;
		\State $\bm {\hat v}=\bm{\mu_{\hat{v}}}+\bm{\sigma_{\hat{v}}}\odot \bm{\hat \epsilon}$;\label{a2:decodeE}
		\State $\bm{\hat y}=sigmoid(\bm{W}^{y}\bm{\mu_z}+\bm{b}^{y})$;\label{a2:whileInB}
		\State update $\bm{AttL}, \{\bm{W_i^{E}}\}_{i=1}^{\chi{+1}}, \{\bm{b_i^{E}}\}_{i=1}^{\chi{+1}}, \{\bm{W_i^{D}}\}_{i=1}^{\chi{+1}}, \{\bm{b_i^{D}}\}_{i=1}^{\chi{+1}}, \bm{W}^{y}, \bm{b}^{y}$ by optimizing Eq.~(\ref{eq:total});\label{a2:optimize}
        \EndWhile \label{a2:whileE}
		\State\Return $\bm{AttL},\{\bm{W_i^{E}}\}_{i=1}^{\chi}, \{\bm{b_i^{E}}\}_{i=1}^{\chi}, \bm{W}^{y}, \bm{b}^{y}$.}\label{a2:return}
	\end{algorithmic}	
\end{algorithm}

\noindent{\bf Algorithm Analysis.}
{The pseudocode of the feature aggregation module is presented in Algorithm~\ref{alg:fa}. First, we randomly initialize the weight and bias matrices for neural layers in line~\ref{a2:initialization}, and produce the initial input vector $\bm{v}$ via the Attention Layer by combining the prosodic vector $\bm{v}_r$ and semantic vector $\bm{v}_s$ in line~\ref{a2:input}; then, we capture the most representative vector for $\bm{v}$ through VAE network in manner of alternately accomplishing encoding and decoding processes in the while loop: the latent vector $\bm{z}$ is constructed through the densely-connected encoder layers in lines [\ref{a2:encodeB}-\ref{a2:encodeE}], and subsequently fed into the densely-connected decoder layers in lines [\ref{a2:decodeB}-\ref{a2:decodeE}], for sake of generating the representation vector $\bm{v}$ that is further optimized along with the latent vector $\bm{\hat y}$ in lines [\ref{a2:whileInB}-\ref{a2:optimize}]; finally, we obtain the hyper-parameter settings of the INPUT and ENCODER stages in line~\ref{a2:return}.} We extract $\bm{\mu_{z}}$ as the target vector $\bm{v}_{t}$~(i.e., $\bm{v_t}:=\bm{\mu_{z}}$). Hence, a unified representation of a rap song, which involves both prosodic information and semantic information, can be generated by repeating lines [\ref{a2:input}-\ref{a2:target}] with the returned hyper-parameters in line~\ref{a2:return}.
\par
{In terms of time and space complexities, they mainly lie on the while loop in lines [\ref{a2:whileB}-\ref{a2:whileE}]. Suppose the convergence state is satisfied when the predefined iteration time $t$ is reached, and the average degree of the network layer is $d$. Then, the time costs for the encoding processing and the decoding processing are both {$\mathcal{O}(d^2\chi tN)$}. Thus, the total time complexity of Algorithm~\ref{alg:fa} is approximately equal to $\mathcal{O}(d^2\chi tN)$. The space complexity is related to $d$ and the number of layer in the network, and is approximate to {$\mathcal{O}(d^2\chi)$}}.
\par
\section{EXPERIMENTS AND RESULTS} 
\label{exp}
The performance of our model~(HAVAE) is evaluated with NextLine prediction, DopeLyrics generation, and rap genre classification tasks by using various metrics. Considering the reality that few online sources provide rap lyrics, we obtain a corpus of rap lyrics sourced from the Internet\footnote{http://ohhla.com/.}. The corpus involves 65,730 songs from 3,154 rappers\footnote{The source code and dataset can be viewed at https://github.com/q9s5c1/HAVAE.}.\par
Table~\ref{tab:song example} shows an example song in the corpus. Each rap song includes four properties: \textbf{Title}~(title of the song), \textbf{Rapper}~(the owner of the song), \textbf{Headings}~(extra information of the song), and \textbf{Lyrics}~(the rap lyrics grouped by song structures). Normally, a rap lyric consists of four primary components, including intro, hook/chorus, verse, and outro. Among these components, verse is the most artistic part of rap songs through which rappers show off their skills, intellect, and rhyming abilities, as well as develop their ideas of songs in depth~\cite{edwards2012rap}.
\begin{table}[htbp]
	\setlength{\belowrulesep}{0pt}
	\setlength{\aboverulesep}{0pt}
	\centering
	\caption{An example rap song}
	\begin{tabular}{ll}
		\toprule
		\multicolumn{1}{l|}{\textbf{Title}} & \textbf{Rapper} \\
		\multicolumn{1}{l|}{Remember the Name} & Fort Minor \\
		\midrule
		\multicolumn{2}{l}{\textbf{Headings}} \\
		Artist & Fort Minor f/ Styles of Beyond \\
		Album & Rising Tied \\
		Typed by & xxxx@xxx.com* \\
		\midrule
		\multicolumn{2}{l}{\textbf{Lyrics}} \\
		\multirow{3}[1]{*}{[\textit{Intro}]} & You ready?! Lets go! \\
		& Yeah, for those of you that want to know what we're all about \\
		& It's like this y'all (c'mon!) \\
		\\
		\multirow{4}[1]{*}{[\textit{Chorus}]} & This is ten percent luck, twenty percent skill \\
		& Fifteen percent concentrated power of will \\
		& Five percent pleasure, fifty percent pain \\
		& And a hundred percent reason to remember the name!\\
		\\
		......&.......\\
		\\
		\multirow{8}[1]{*}{[\textit{Verse}]} & Forget Mike - Nobody really knows how or why he works so hard \\
		& It seems like he's never got time \\
		& Because he writes every note and he writes every line \\
		& And I've seen him at work when that light goes on in his mind\\
		& It's like a design is written in his head every time\\
		& Before he even touches a key or speaks in a rhyme\\
		& And those motherfuckers he runs with, those kids that he signed?\\
		& Ridiculous, without even trying, how do they do it?!\\
		\\
		\multirow{3}[1]{*}{[\textit{Outro}]} & Yeah! Fort Minor \\
		& M. Shinoda - Styles of Beyond \\
		& Ryu! Takbir! Machine Shop! \\
		\bottomrule
		\multicolumn{2}{l}{*\scriptsize We only display the format of this value for the sake of privacy.}\\
	\end{tabular}%
	\label{tab:song example}%
\end{table}%
\vspace*{-2em}
\subsection{NextLine Prediction Task}
The NextLine prediction task is developed in~\cite{Malmi2016dopelearning}. Given a rap song with a sequence of $k$ rap lines, we assume that the first $\kappa~(\kappa<k)$ lines, denoted by $\bm Q=\{s_i\}_{i=1}^{\kappa}$, are available. Let $\bm C={\left\{c_i\right\}}_{i=1}^{\pi}$ be a set of candidate lines. The task is to predict the subsequent ${(\kappa+1)}^{th}$ line from $\bm C$(i.e., $s_{\kappa+1}$). The closest rap line $c_i \in \bm C$ is selected as the matching object.\\

\noindent{\bf Dataset.}
\noindent Following~\cite{Malmi2016dopelearning}, we extract the dominant parts~(i.e., verses) of rap songs and obtain 16,697 verses in total\footnote{A number of rap songs in the tested corpus are lack of component labels, and it is difficult to extract verses from these songs without professional knowledge. To eliminate the bias incurred by non-expert labeling and make fair comparisons, the songs with explicit “verse” labels are extracted as the experimental dataset.}. The verses are divided into lines to obtain a dataset of 810,567 lines. Then, we separate the dataset by song to obtain the train set (50\%), validation set (25\%), and test set (25\%). \\

\noindent{\bf Baseline Methods.} We compare the proposed HAVAE with the following methods:
\begin{itemize}
	\item {\bf EndRhyme}~\cite{Malmi2016dopelearning}, which considers the number of matching vowel phonemes at the end of candidate line $c_i$ and $s_\kappa$;
	\item {\bf rhyme2vec}, our novel rhyme embedding method, as described in Section~\ref{sec:fe};
	\item {\bf NN5}~\cite{Malmi2016dopelearning}, a character-level neural network for rap line encoding, which takes five previous lines as the query (i.e.,$\{s_{\kappa-i}\}_{i=0}^{4}$); 
	\item {\bf doc2vec}~\cite{quoc2014distributed}, a popular sentence embedding method, which handles $\{s_\kappa;c_i\}$ as a unified paragraph;
	\item {\bf DopeLearning}~\cite{Malmi2016dopelearning}\footnote{In the current paper, we report the results in the original work, and reproduce it on the crawled dataset.}, a state-of-the-art rap lyric representation learning method, which concatenates  a series of statistical characteristics, including the features of EndRhyme, EndRhyme-1~(number of matching vowel phonemes at the end of $c_i$ and $s_{\kappa-1}$), OtherRhyme~(average number of matching vowel phonemes per word), LineLength~(line similarity of $c_i$ and $s_\kappa$), BOW~(Jaccard similarity between the corresponding bags of words of $c_i$ and $s_\kappa$), BOW5~(Jaccard similarity between the corresponding bags of words of five previous lines and $s_\kappa$), LSA~(latent semantic analysis similarity of $c_i$ and $s_\kappa$), and NN5~(confidence value generated from the last $softmax$ layer);
	\item {\bf early fusion}~\cite{chen2017visual}, a widely used multi-modal aggregation method, which concatenates all of the features as a unified representation~(i.e., $\bm {v_t}:=[\bm{v_r},\bm{v_s}]$);
	\item {\bf EF-AE}, a variant of HAVAE, which adopts the same learning manipulations as that of HAVAE, but bypasses the sampling strategy and renders $[\bm{v_r},\bm{v_s}]$ as the input of the network; and 
	\item {\bf EF-VAE}, another variant of HAVAE, which renders $[\bm{v_r},\bm{v_s}]$ as the input of the VAE network instead of the INPUT stage.  
\end{itemize}
The methods can be categorized into three types on the basis of the involved information, namely, the prosodic, semantic, and both~(prosodic and semantic) types.\\

\noindent{\bf Experiment Setting.}
In the training phase, every query line $s_\kappa$ has two candidate lines. One line is the true next line of the given line~(ground truth) and serves as a positive example. The other line is a randomly selected line from the corpus, and serves as a negative example. In the testing phase, each query line has a candidate set that contains the true next line and the randomly chosen 299 lines.\par
We combine each query line with a candidate line to obtain a pair. In contrast to BOW5 and NN5 that consider the five previous lines when semantic features are extracted, we only utilize $s_\kappa$ for the materials. 
Following the method in~\cite{Malmi2016dopelearning}, the representation obtained by the DopeLearning method is fed into the SVM$^{rank}$~\cite{thorsten2006training} tool to generate the ranking. In this scenario, a complicated issue is observed in which the representations derived by the other methods become highly dimensional; thus, the SVM$^{rank}$ tool is infeasible. To resolve this issue, we design the Rank Layer, expressed as $score=sigmoid(W\bm{v_t}+b)$, where $score\in [0,1]$ is a confidence value that represents the relevance of the candidate and query lines. The higher the $score$, the better the performance. \\

\noindent{\bf Evaluation Results. } 
The performance of the approaches is compared and evaluated by using the metrics of mean rank, mean reciprocal rank (MRR), and recall at $*$ ($*$=1,5,30,150) denoted as Rec@$*$. Let $rank_i$ denote the rank position of the correct next line in the candidate set for the $i$-${th}$ query. Mean rank is the average value of $\sum_{i=1}^{N}rank_i$, and a lower value is better~(i.e., the value of the mean rank falls between $1$ and $300$). MRR is the average score of $\sum_{i=1}^{N}\frac{1}{rank_i}$, and a higher value is better. Rec@$*$ is the probability of ${rank_i} {\leq} *$, and higher values are better.
\par

Table~\ref{table:results} presents the experimental results. The proposed HAVAE is superior to the state-of-the-art methods. The following observations are also derived:

\begin{table}[t]
	\setlength{\belowrulesep}{0pt}
	\setlength{\aboverulesep}{0pt}
	\centering
	\caption{Results of the proposed method and the baseline methods}
	\scalebox{0.8}{
		\begin{tabular}{p{3em}|l|c|c|c|c|c|c}
			\toprule
			\multicolumn{1}{l|}{Information} & Methods & Mean rank & MRR   & Rec@1 & Rec@5 & Rec@30 & Rec@150 \\
			\midrule
			\multicolumn{1}{l|}{\multirow{2}[0]{*}{prosodic}} & EndRhyme & 103.2* & 0.140* & 0.077* & 0.181* & 0.344* & 0.480* \\
			\cmidrule{2-8}
			& rhyme2vec &17.7&0.463&0.347&0.592 &0.841&0.981  \\			
			\midrule
			\multicolumn{1}{l|}{\multirow{2}[0]{*}{semantic}} & NN5& 84.7* & 0.067* & 0.020* & 0.083* & 0.319* & 0.793* \\
			\cmidrule{2-8}
			& doc2vec &15.5&0.430&0.293&0.588&0.870&0.985\\
			\midrule
			\multirow{6}[0]{*}{both} & DopeLearning & 60.8*/79.9 & 0.243*/0.168 & 0.169*/0.102 & 0.304*/0.220 & 0.527*/0.446 & 0.855*/0.775 \\
			\cmidrule{2-8}
			& early fusion &9.6&0.588&0.464&0.738&0.926&0.991\\
			\cmidrule{2-8}
			& EF-AE &5.3&0.771&0.683&0.879&0.966&0.995\\
			\cmidrule{2-8}
			& EF-VAE &2.3&0.941&0.914&0.973&0.990&0.997\\
			\cmidrule{2-8}
			& \textbf{HAVAE} &\textbf{1.2}&\textbf{0.982}&\textbf{0.973}&\textbf{0.993}&\textbf{0.999}&\textbf{1.000}\\			
			\bottomrule
			\multicolumn{8}{l}{ * denotes the reported results from the original paper~\cite{Malmi2016dopelearning}.} \\
			
		\end{tabular}%
	}
	\label{table:results}%
\end{table}%

\begin{itemize}
\item \textit{Effectiveness of HAVAE}. Our proposed HAVAE achieved superior performance compared with the other evaluated methods. The mean rank is 1.2~(extremely close to 1), which proves that HAVAE outperformed the most advanced approaches by a substantial margin. The mean rank of DopeLearning is 60.8 in the original paper, and 79.9 by using the settings in our study, and both values are higher than 60. The MRR of HAVAE is 0.982, and outperforms DopeLearning by $73.9^*/81.4$ percentage points. Furthermore, the evaluated performance of Rec@$*$ suggests a similar observation, i.e., HAVAE can yield better results. In summary, HAVAE exceeded the state-of-the-art approaches on the various evaluation metrics. HAVAE utilized an effective technology by combining fine-grained prosodic and semantic features~(distributed representation learning instead of simple statistical characteristics) and by executing an appropriate feature aggregation strategy~(attentive VAE-based framework instead of easy early fusion).
\item \textit{Effectiveness of rhyme2vec}. Compared with EndRhyme, rhyme2vec has more promising performance on the basis of the evaluation metrics~(i.e., mean rank, MRR, and Rec@$*$). This outcome demonstrates that the prosodic vector representation learned by our rhyme embedding method, which comprehensively considers various rhyme schemes, is much more effective than the state-of-the-art prosodic features-based approach.\par
\item \textit{Effectiveness of the prosodic-enhanced representation}. When the semantic representations~(doc2vec) or prosodic representations~(rhyme2vec) are independently utilized, the mean rank is higher than 15 and the MRR score is lower than 0.5. The performance is significantly improved after fusing both representations. Similarly, DopeLearning performs better than EndRhyme and NN5. This result demonstrates that combining prosodic and semantic information is more effective than individually rendering prosodic or semantic information. Accordingly, our prosodic-enhanced vector is more representative than the traditional semantic vectors. 
\item \textit{Effectiveness of the feature aggregation module}. The early fusion method renders the concatenation of all features as the unified representation similar to that in DopeLearning. The mean rank and MRR scores are 9.6 and 0.588, respectively. Then, by taking the early fusion approach as the baseline, the EF-AE method is also improved because of the adoption of the autoencoder network that can learn useful information from the inner relationship of the input~\cite{bengio2009learning}. Notably, EF-VAE performs better than EF-AE. This result reveals another notable aspect, that is, VAE is more powerful than AE with additional noises when generating the representations of rap lyrics. In conclusion, the superiority of HAVAE over EF-VAE indicates that the attention mechanism, which amplifies the mutual relationship between semantic and prosodic information, can derive substantial benefits.
\item {\textit{Remarks}}. Doc2vec is comparable to rhyme2vec for some of the evaluation metrics. NN5 works better than EndRhyme in terms of mean rank and Rec@150 scores. The results can be attributed to rappers initially imaging the topic of a song, which is best represented by semantic information. However, different rhymes may be used to realize the idea. For instance, ``mommy[m\textturnscripta m\textbf{\textsci}]'' and ``daddy[d\ae d\textbf{\textsci}]'' can be replaced with ``mother[m\textturnv \textbf{\textipa{D}\textschwa}]'' and ``father[fa: \textbf{\textipa{D}\textschwa}]'' without changing the  fundamental semantics.
\end{itemize}
 \begin{figure}[t]
	\centering
		\includegraphics[width=0.88\textwidth]{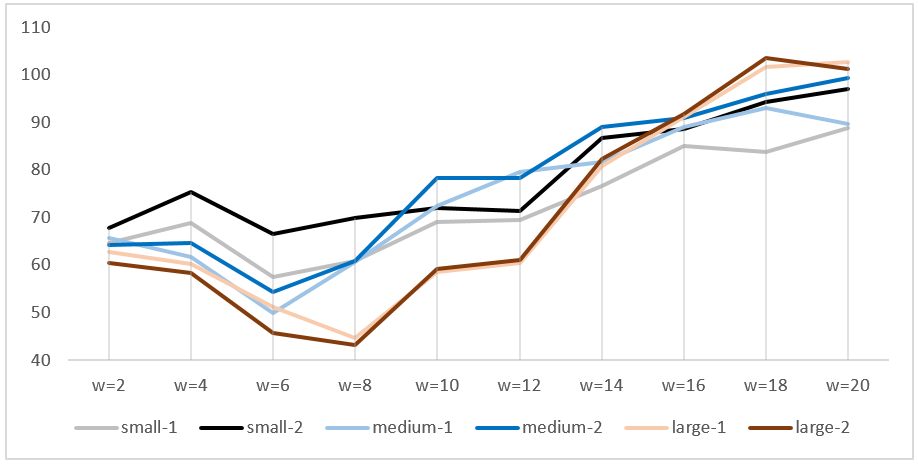}
\caption{Performance~(mean rank value) with different sliding window size and corpora}
	\label{fig:window}
\end{figure}
\noindent{\bf {Auxiliary Results.}}To comprehensively verify the performance of HAVAE, we implement several additional experiments. We extract a small subset from the original corpus by randomly selecting 20,000 samples from the train set and 1,000 samples from the test set. We investigate the performance of HAVAE when changing the sliding window size, adopting different rhyme schemes, varying the context size, and tuning the balance parameter $\alpha$ in Eq.~(\ref{eq:total}). Each experimental test is evaluated 10 times and the average values are reported.\par

Unlike semantic information, which comes from neighbor words or context, prosodic information may exist in words that are not close to each other. Therefore, the sliding window size $w$ of the rhyme2vec strategy should be large enough to capture the prosodic information, which can cross multiple lines with little noise embedded. To evaluate the effect of $w$, we train the rhyme embedding approach with various values. Moreover, we extract five different corpora as follows
\begin{itemize}
	\item \textbf{small-1}, which involves 3,339~(20\%) songs randomly selected from the entire corpus;
	\item \textbf{small-2}, ditto;
	\item \textbf{medium-1}, which involves 6,678~(40\%)  songs randomly selected from the entire corpus;
	\item \textbf{medium-2}, ditto;
	\item \textbf{large-1}, which involves 10,018~(60\%)  songs randomly selected from the entire corpus;
	\item \textbf{large-2}, ditto.
\end{itemize}
The performance on mean rank scores is presented in Fig.~\ref{fig:window}, where the window size $w$ varies from 2 to 20 with an interval of 2. We observe that the larger the corpus, the more stable the performance of different $w$. Specifically, the average difference of mean rank value between \textbf{small-1} and \textbf{small-2} is 7.3, while it is 3.2 between \textbf{medium-1} and \textbf{medium-2}. The smallest one is 0.7 between \textbf{large-1} and \textbf{large-2}. Another observation is that the larger the corpus, the better the performance of different $w$. Specifically, the best result of large corpora is 43.3, while the mean rank score of small corpora is 57.5. We can infer that the sensitivity of $w$ consists in the distribution of training corpus, i.e., large corpora have stable distributions and can capture more prosodic features than small ones. Moreover, it is not the larger of $w$ the better. For example, when training on \textbf{large-1} and \textbf{large-2}, the performance of $w>12$ is worse than that of $w\leq 12$. The reason could be that too large $w$ brings into too much irrelevant phonemes which decrease the quality of rhyme learning procedure. 
\par


\begin{figure}
	\begin{minipage}[b]{0.5\linewidth}
				
		\centering
		\renewcommand{\arraystretch}{1.5}
		\captionof{table}{{Performance with different rhyme schemes}}
		\setlength{\belowrulesep}{0pt}
		\setlength{\aboverulesep}{0pt}
		\begin{tabular}{c|c|c}
			\hline	
			
			\rule{0pt}{12pt}		
			\textbf{Rhyme scheme}& \textbf{Mean rank} & \textbf{MRR} \\[2pt]\hline
			\rule{0pt}{12pt}
			C-Line & 60.1  & 0.154 \\
			\midrule
			Skip-Line & 61.3  &   0.150\\
			\midrule
			EF-CS & 24.1  &   0.407\\
			\midrule
			\textbf{rhyme2vec}  &\textbf{17.7} & \textbf{0.463} \\
			[1pt]
			\hline		
		\end{tabular}
		\vspace*{-0.5em}
		
		\label{table:rs}
	\end{minipage}
	\begin{minipage}[b]{0.5\linewidth}
		\centering
		\centering
		\includegraphics[width=0.8\textwidth]{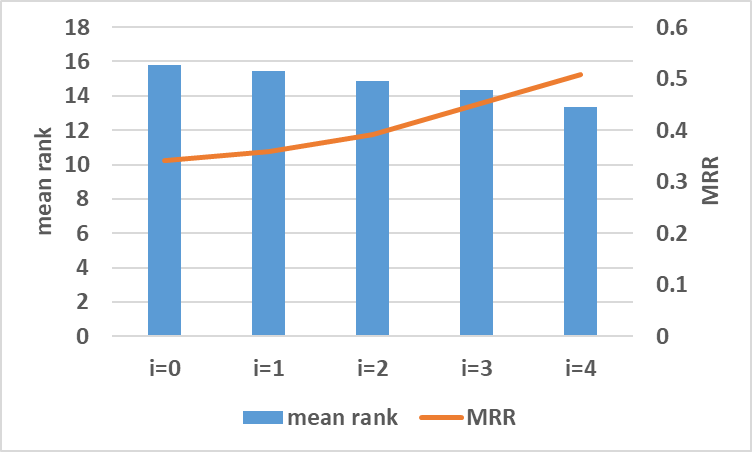}
		\vspace*{-0.5em}
		\caption{Performance with different context size}
		\label{fig:i}
	\end{minipage}
\end{figure}
Second, we conduct experiments w.r.t. the performance of different rhyme schemes in the NextLine Prediction task, and the results are shown in Table~\ref{table:rs}. To simplify the presented data, we use mean rank and MRR as the evaluation metrics. ``C-Line'' constructs the prosodic representation for monorhyme, while ``Skip-Line'' produces the prosodic representation for alternative rhymes. ``EF-CS'' utilizes the linear concatenation of the vectors generated by C-Line and Skip-Line models as the prosodic representation. ``rhyme2vec'' uses the prosodic representation generated by the newly proposed rhyme2vec method that integrates both monorhyme and alternative rhymes through hierarchical attention mechanism. We can observe from the table that rhyme2vec and EF-CS beat C-Line and Skip-Line by a substantial margin, which attribute to their comprehensive considerations of monorhyme and alternative rhymes. Moreover, rhyme2vec is superior to EF-CS, due to its effective strategy to appropriately fusing two kinds of rhymes.

Third, it was assumed in~\cite{Malmi2016dopelearning} that rap lyric learning performance could be improved by changing the learning context from one previous line~($s_\kappa$) to multiple prevous lines~(five previous lines, $\{s_{\kappa-i}\}_{i=0}^4$). To test this assumption, we assess the performance of HAVAE in five context conditions with different number of previous lines. 
Concretely, the context size increases from 1 to 5, and we use $i=*$ to render $i:=\{0,...,*\}$ as the short version, as shown in Fig. 4. We notice that the performance of HAVAE is improved to a limited extent when the context size increases, while this rising trend can be remarkably observed merely when the learning context is expanded to a relatively large size. However, given that the average number of words of a single line is $\phi$, an added line takes additional $\mathcal{O}(\phi^2)$ time and $\mathcal{O}(\phi N)$ space cost. To make the comparisons fair and convincible, we illustrate the performance of HAVAE as $i=0$ throughout our experiments. HAVAE yields excellent results even with this non-predominant parameter configuration, as shown in Table~\ref{table:results}.\par
\begin{figure}[t]
	
	\centering
	\begin{minipage}[t]{0.43\textwidth}  
		\centering
		\includegraphics[width=5.3cm]{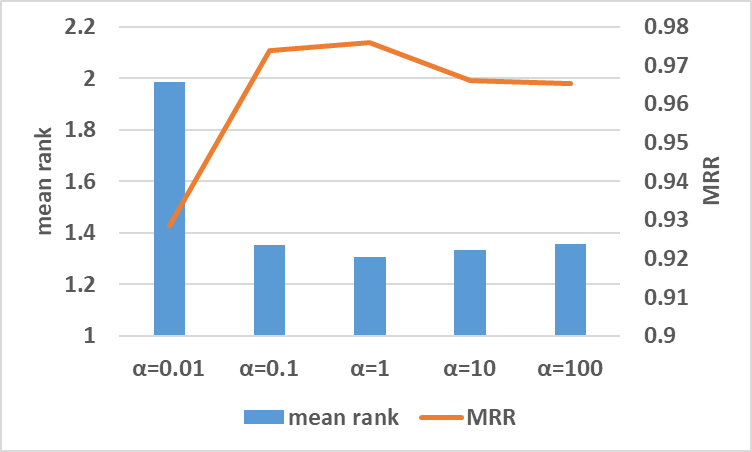}
		\caption{Mean rank and MRR with different $\alpha$
		}
		\label{fig:alpha}
	\end{minipage}
	\qquad
	\begin{minipage}[t]{0.43\textwidth}  
		\centering
		\includegraphics[width=5.3cm]{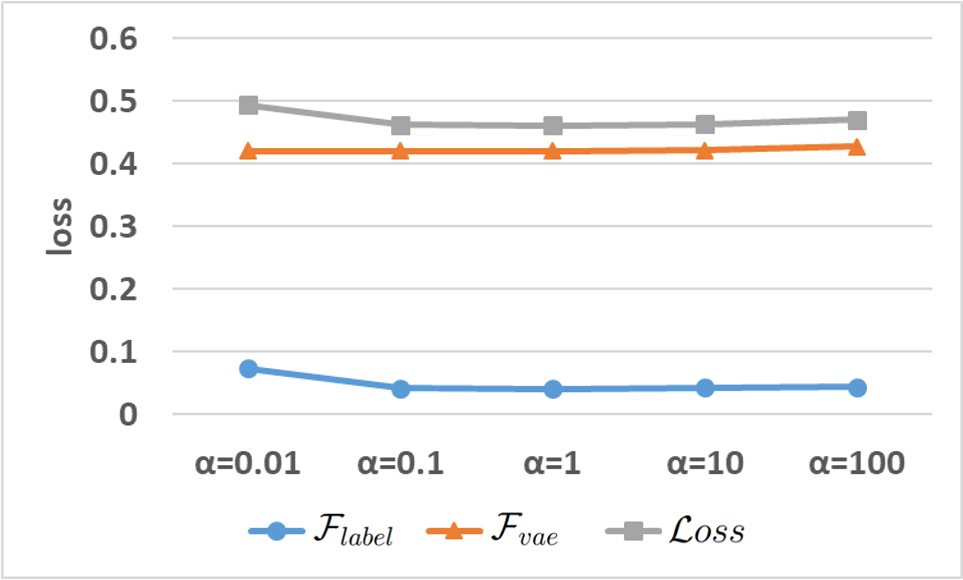}
		\caption{Loss sensitivity with different $\alpha$}
		\label{fig:loss}
	\end{minipage}
\end{figure}
Finally, we vary $\alpha$ to test its effects on the performance of HAVAE. The configuration of $\alpha$ is crucial for constructing an effective VAE network through which HAVAE accomplishes its core manipulation, that is, aggregating diverse features extracted from rap lyrics in different views. 
According to Eq.~(\ref{eq:total}), the performance of HAVAE is completely determined by the effectiveness of the attentive VAE-based network when $\alpha = 0$. More emphasis is laid on the label information when increasing $\alpha$. As shown in Fig.~\ref{fig:alpha}, the best performance is achieved when $\alpha=1$. This finding reveals that $\mathcal{F}_{vae}$ and $\mathcal{F}_{label}$ are both essential for training HAVAE, and thus, both feature aggregation information and label information are important for the rap lyric analysis. Fig.~\ref{fig:loss} exhibits the variations of different kinds of losses, including $\mathcal{F}_{vae}$, $\mathcal{F}_{label}$, and their combination, denoted as $\mathcal{L}oss$, which describes the entire loss of feature aggregation in HAVAE. A similar conclusion to that drawn from Fig.~\ref{fig:alpha} can be derived, that is, a comprehensive consideration of both feature aggregation information and label information contributes to effectively optimizing the VAE network of HAVAE and eventually, facilitates constructing most representative vectors for rap lyrics.

\begin{table}[t]
	\caption{An example of generated DopeLyrics}
	\begin{center}
			\begin{tabular}{l>{\raggedleft\arraybackslash}p{0.42\columnwidth}}  
				\toprule
				\textbf{Rap Lyrics}& \textbf{Reference}\\
				\midrule
				curse ya breath, there's no person left & [Precision : 7L \& Esoteric]\\
Got your BM in the trap she say y'all losers over there & [Keep Gettin' Money : Fredo Santana]\\
I'm a draw down on you and your partner so perfect & [Here I Go : Hustle Gang]\\
and stay fly from her toes to her hair & [Slow Down : Lil' O]\\
I once heard don't forget & [My Life : MC Eiht]\\
Reckless youngsters, no direction & [Temptation : Dizzee Rascal]\\
And everyday alive is just another closer to death & [Black Winter Day : Jedi Mind Trick]\\
In this life you need to know there's no perfection & [Why Must Life be This Way : Califa Thugs]\\
To free us from our lower selves & [Absolute Truth : Flame]\\
Don't point a finger over herre & [Sick of Being Lonely : Field Mob]\\
It's Murder! Irv don't hurt 'em & [Girls All Around the World : Lloyd]\\
OK, I can feel me gettin closer to them & [Treddin' On Thin Ice : Wiley]\\
Hammered it across his cranium, he fell over dead & [Basehead Attack : Insane Clown Posse]\\
She cryin, headshots put her to rest & [Niggas Bleed : Notorious B.I.G]\\
like fresh over fresh & [Pistols Blazin :  3X Krazy]\\
				\bottomrule			
			\end{tabular}
	\end{center}	
	\label{generated}
\end{table}
 \begin{figure}[t]
	\centering
	\includegraphics[width=0.8\linewidth]{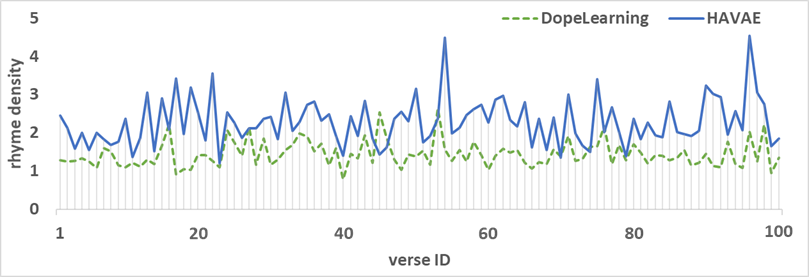}
	\caption{{Rhyme density scores of generated verses}}
	\label{fig:rd2016}
\end{figure}
\subsection{{DopeLyrics Generation Task}} 
The DopeLyrics generation task is a derivation of the NextLine prediction task. Table~\ref{generated} presents an example of DopeLyrics. The ``generated'' verse starts with a randomly selected rap line, while the second line is the most relevant line to the first one. The same treatment is applied to sequential lines. The generated verses are fixed to 16 lines~\cite{Malmi2016dopelearning}. Unlike in traditional natural language tasks, a new set of literatures is generated from the existing rap songs. We call this generation task as DopeLyrics generation task.\\


\noindent {\bf Dataset and Setting.} 
	A previous work~\cite{Malmi2016dopelearning} generated $100$ verses by using $100$ random rap lines with the  DopeLearning approach. In this work, we use the same $100$ first rap lines, and generate $100$ verses. The subsequent lines are selected from the corpus that has been created by HAVAE in the NextLine prediction task.\\

\noindent {\bf Evaluation and Results.} 
	A good verse is fluent, contains an explicit motif, and possesses a catchy rhyme. However, quantitatively evaluating the first two factors is complicated, and extremely costly in non-mechanical ways. Therefore, we only evaluate the rhyme quality of the generated verses. Rhyme density~\cite{Malmi2016dopelearning} is a measure for examining the prosodic quality of rap lyrics technically, and is computed as the average length of the longest rhyme~(i.e., the longest matching vowel sequence) per word in a rap song.\par
	


Fig.~\ref{fig:rd2016} illustrates the rhyme density scores of the generated verses.  As depicted in Fig.~\ref{fig:rd2016}, HAVAE performs better than DopeLearning with average rhyme densities of 2.278 versus 1.436. These evaluation results indicate that the verses generated by HAVAE possess  higher rhyme densities than those generated by DopeLearning. In other words, the rap lyrics produced by HAVAE are more rhyming.  

\subsection{Rap Genre Classification}
To demonstrate the generalizability of our method, we design a rap genre classification task to complement the song level. This task is defined as follows:
\begin{mydef}
	Consider a set of rap songs $\mathcal{S}=\{S_i\}_{i=1}^{N}$ and a set of genre labels $\mathcal{G}=\{G_i\}_{i=1}^{\pi}$. The aim is to predict a proper set of labels $ Y \subseteq  G$ for each song $S_i$. 
\end{mydef}
In essence, this experiment is a multi-label classification task.\\

\noindent {\bf Dataset.} Considering prevailing copyright issue and the absence of comprehensive research on rap genre classification, we create a dataset from the corpus, and it consists of 10,167 songs from nine mainstream rap genres~(Alternative, Christian, East Coast, Grime, Hardcore, Horrorcore, Midwest, Southern, and West Coast). These genres are used as the labels of the songs.\\

\noindent {\bf Baseline Models.} The baseline methods include the following:
\begin{itemize}
\item {\bf RhymeAPP}~\cite{hirjee2010rhyme}, a lyrical analysis tool considering the statistical features of rap songs; 
\item {\bf rhyme2vec}, is our novel rhyme embedding approach and includes prosodic information; 
\item {\bf doc2vec}, which includes semantic information; and 
\item {\bf HAN}~\cite{Tsaptsinos2017lyrics}, a state-of-the-art music genre classification approach for intact lyrics.
\end{itemize}\par

\noindent {\bf Experiment Setting.} 
We feed features generated by comparison models into the classification network to obtain the genre prediction vector $\bm y$ for each song and subsequently realize the probability of tagging the nine labels to a given song. We also construct threshold $t$ to determine whether the given song can be categorized into the corresponding genre. If $ {y_i}\leq t$, $y'_i=0,$ otherwise $y'_i=1$. Here $ {y_i}$ is the $i$-${th}$ element of $\bm y$. We use $\bm{y'}=\{y'_i\}_{i=1}^9$ as the predicted label set. Specifically, we attach the $i$-${th}$ genre label with a rap song only when $y'_i=1$.\\
\begin{figure}[t]
	\begin{minipage}[t]{0.5\linewidth}  
		\centering
		\includegraphics[width=6cm]{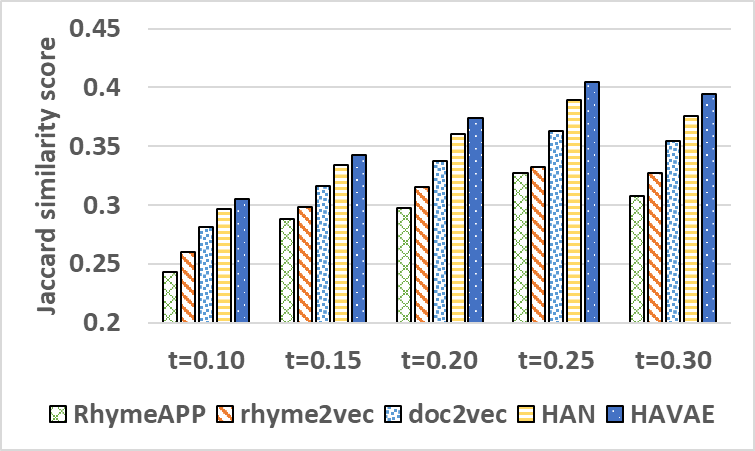}
		\caption{{Jaccard similarity score}}
		\label{fig:jaccard}
	\end{minipage}
	\begin{minipage}[t]{0.5\linewidth}  
		\centering
		\includegraphics[width=6cm]{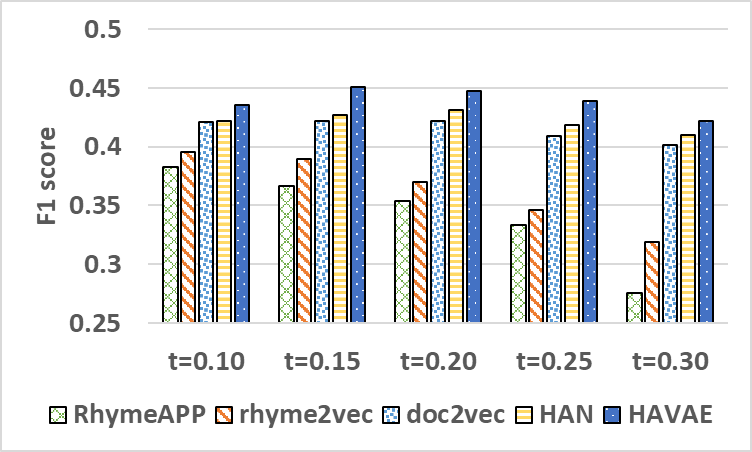}
		\caption{{F1 score} }
		\label{fig:f1}
	\end{minipage}	
\end{figure}

\noindent {\bf Evaluation and Results.} To facilitate comparisons between our method and baselines, we use Jaccard similarity score~\cite{Real1996The} and F1 score as evaluation metrics. All of the compared approaches perform rap genre classifications for ten times, and their average performance in terms of Jaccard similarity score and F1 score are shown in Fig.~\ref{fig:jaccard} and Fig.~\ref{fig:f1}, respectively.\par
Overall, HAVAE dominates the baseline techniques under both metrics, and especially it performs the best when $t$=0.25 under the metric of Jaccard similarity (i.e., 0.405) and $t$=0.15 under the metric of F1 (i.e., 0.450). These observations validate the outstanding performance of HAVAE in rap music classification task. Furthermore, we notice that rhyme2vec outperforms RhymeAPP~\cite{hirjee2010rhyme}, which suggests that rhyme2vec is a competitive approach to extract prosodic information of rap lyrics at song-level in addition to line-level. This is coincident with the observations
reported in the NextLine prediction and DopeLyrics generation tasks. The superiority of HAVAE over HAN~\cite{Tsaptsinos2017lyrics} indicates that an excellent ability of discriminating rap music genres can be possessed from a unified perspective with both semantic and prosodic analyses.\par
Generally speaking, HAVAE can be expected to approach a variety of music learning issues besides lyrics prediction, lyrics generation, and music classification, due to its superior ability of capturing prosodic and semantic information simultaneously. \par
\section{CONCLUSIONS AND FUTURE WORK}
We proposed a general framework to learn rap lyrics~(i.e., HAVAE), and we considered both semantic and prosodic features. Several effective strategies were introduced, such as, rhyme2vec, VAE-based feature aggregation module, and hierarchical attention mechanism. The effectiveness of the framework was experimentally demonstrated by the compared results with state-of-the-art approaches on benchmark datasets in three types of tasks.\par

In the future, we plan to examine a wider range of information embedded in rap lyrics besides its prosodic and semantic information, and design other effective techniques to extract representative vectors for this information. Furthermore, we will evaluate the proposed framework for common tasks~(music recommendation systems) and other types of prosodic text~(poetry).
\vspace{-1em}
\begin{acknowledgements}
	This work was supported in part by the National Natural Science Foundation of China under Grant No.U1636116, 11431006, 61772278, the Research Fund for International Young Scientists under Grant No. 61650110510 and 61750110530, the Ministry of education of Humanities and Social Science project under grant 16YJC790123, and Open Fundation for Key Laboratory of Information Processing and Intelligent Control in Fujian Province under Grant MJUKF201705.
\end{acknowledgements}

\vspace{-2em}
\bibliographystyle{spbasic}      
\bibliography{w3j2018}

\begin{thebibliography}{38}
\providecommand{\natexlab}[1]{#1}
\providecommand{\url}[1]{{#1}}
\providecommand{\urlprefix}{URL }
\expandafter\ifx\csname urlstyle\endcsname\relax
  \providecommand{\doi}[1]{DOI~\discretionary{}{}{}#1}\else
  \providecommand{\doi}{DOI~\discretionary{}{}{}\begingroup
  \urlstyle{rm}\Url}\fi
\providecommand{\eprint}[2][]{\url{#2}}

\bibitem[{Addanki and Wu(2013)}]{Addanki2013Unsupervised}
Addanki K, Wu D (2013) Unsupervised rhyme scheme identification in hip hop
  lyrics using hidden {Markov} models. In: Proceedings of the International
  Conference on Statistical Language and Speech Processing, pp 39--50

\bibitem[{Alexey and Ivan(2017)}]{alexey2017music}
Alexey T, Ivan PY (2017) Music generation with variational recurrent
  autoencoder supported by history. arXiv preprint arXiv:170505458

\bibitem[{Association(1999)}]{international1999handbook}
Association IP (1999) Handbook of the International Phonetic Association: A
  guide to the use of the International Phonetic Alphabet. Cambridge University
  Press

\bibitem[{Bengio(2009)}]{bengio2009learning}
Bengio Y (2009) Learning deep architectures for {AI}. Machine Learning
  2(1):1--127

\bibitem[{Bryant et~al(1990)Bryant, MacLean, Bradley, and
  Crossland}]{bryant1990rhyme}
Bryant PE, MacLean M, Bradley LL, Crossland J (1990) Rhyme and alliteration,
  phoneme detection, and learning to read. Developmental psychology 26(3)

\bibitem[{Chen et~al(2017)Chen, Wang, and Liu}]{chen2017visual}
Chen X, Wang Y, Liu Q (2017) Visual and textual sentiment analysis using deep
  fusion convolutional neural networks. In: Proceedings of the 2017 IEEE
  International Conference on Image Processing, pp 1557--1561

\bibitem[{Dosovitskiy and Brox(2016)}]{Dosovitskiy2016Generating}
Dosovitskiy A, Brox T (2016) Generating images with perceptual similarity
  metrics based on deep networks. In: Proceedings of the 30th International
  Conference on Neural Information Processing Systems, pp 658--666

\bibitem[{Duddington(2012)}]{duddington2012espeak}
Duddington J (2012) {eSpeak} text to speech.
  \urlprefix\url{http://espeak.sourceforge.net}

\bibitem[{Edwards(2012)}]{edwards2012rap}
Edwards P (2012) How to Rap. Random House

\bibitem[{Fabius and van Amersfoort(2014)}]{fabius2014variational}
Fabius O, van Amersfoort JR (2014) Variational recurrent auto-encoders. arXiv
  preprint arXiv:14126581

\bibitem[{Hadjeres et~al(2017)Hadjeres, Nielsen, and Pachet}]{hadjeres2017glsr}
Hadjeres G, Nielsen F, Pachet F (2017) {GLSR-VAE}: Geodesic latent space
  regularization for variational autoencoder architectures. arXiv preprint
  arXiv:170704588

\bibitem[{He et~al(2012)He, Zhou, and Jiang}]{he2012generating}
He J, Zhou M, Jiang L (2012) Generating chinese classical poems with
  statistical machine translation models. In: Proceedings of the 26th AAAI
  Conference on Artificial Intelligence, pp 1650--1656

\bibitem[{Hirjee and Brown(2009)}]{hirjee2009automatic}
Hirjee H, Brown DG (2009) Automatic detection of internal and imperfect rhymes
  in rap lyrics. In: Proceedings of the International Society for Music
  Information Retrieval Conference, pp 711--716

\bibitem[{Hirjee and Brown(2010{\natexlab{a}})}]{hirjee2010rhyme}
Hirjee H, Brown DG (2010{\natexlab{a}}) Rhyme analyzer: An analysis tool for
  rap lyrics. In: Proceedings of the 11th International Society for Music
  Information Retrieval Conference

\bibitem[{Hirjee and Brown(2010{\natexlab{b}})}]{Hirjee2010Using}
Hirjee H, Brown DG (2010{\natexlab{b}}) Using automated rhyme detection to
  characterize rhyming style in rap music. Empirical Musicology Review
  5(4):121--145

\bibitem[{Hou et~al(2017)Hou, Shen, Sun, and Qiu}]{hou2017deep}
Hou X, Shen L, Sun K, Qiu G (2017) Deep feature consistent variational
  autoencoder. In: Proceedings of the IEEE Winter Conference on Applications of
  Computer Vision, pp 1133--1141

\bibitem[{Hu et~al(2017)Hu, Bai, Cheng, Deng, Guo, Hu, Krishnan, and
  Wang}]{hu2017medj}
Hu X, Bai K, Cheng J, Deng Jq, Guo Y, Hu B, Krishnan AS, Wang F (2017) {MeDJ}:
  multidimensional emotion-aware music delivery for adolescent. In: Proceedings
  of the 26th International Conference on World Wide Web Companion, pp 793--794

\bibitem[{Kingma and Welling(2013)}]{kingma2013auto}
Kingma DP, Welling M (2013) Auto-encoding variational bayes. arXiv preprint
  arXiv:13126114

\bibitem[{Kingma et~al(2014)Kingma, Rezende, Mohamed, and
  Welling}]{Kingma2014Semi}
Kingma DP, Rezende DJ, Mohamed S, Welling M (2014) Semi-supervised learning
  with deep generative models. In: Proceedings of the 27th International
  Conference on Neural Information Processing Systems, pp 3581--3589

\bibitem[{Lei et~al(2017)Lei, Wang, Liu, Ilievski, He, and Kan}]{lei2017swim}
Lei W, Wang X, Liu M, Ilievski I, He X, Kan MY (2017) Swim: a simple word
  interaction model for implicit discourse relation recognition. In:
  Proceedings of the 26th International Joint Conference on Artificial
  Intelligence, pp 4026--4032

\bibitem[{Lei et~al(2018{\natexlab{a}})Lei, Jin, Kan, Ren, He, and
  Yin}]{lei2018sequicity}
Lei W, Jin X, Kan MY, Ren Z, He X, Yin D (2018{\natexlab{a}}) Sequicity:
  Simplifying task-oriented dialogue systems with single sequence-to-sequence
  architectures. In: Proceedings of the 56th Annual Meeting of the Association
  for Computational Linguistics (Volume 1: Long Papers), pp 1437--1447

\bibitem[{Lei et~al(2018{\natexlab{b}})Lei, Xiang, Wang, Zhong, Liu, and
  Kan}]{lei2018linguistic}
Lei W, Xiang Y, Wang Y, Zhong Q, Liu M, Kan MY (2018{\natexlab{b}}) Linguistic
  properties matter for implicit discourse relation recognition: Combining
  semantic interaction, topic continuity and attribution. In: Proceedings of
  the AAAI Conference on Artificial Intelligence, vol~32

\bibitem[{Lei et~al(2019)Lei, Xu, Aw, Xiang, and Chua}]{lei2019revisit}
Lei W, Xu W, Aw A, Xiang Y, Chua TS (2019) Revisit automatic error detection
  for wrong and missing translation--a supervised approach. In: Proceedings of
  the 2019 Conference on Empirical Methods in Natural Language Processing and
  the 9th International Joint Conference on Natural Language Processing
  (EMNLP-IJCNLP), pp 941--951

\bibitem[{Malmi et~al(2016)Malmi, Takala, Toivonen, Raiko, and
  Gionis}]{Malmi2016dopelearning}
Malmi E, Takala P, Toivonen H, Raiko T, Gionis A (2016) {DopeLearning}: A
  computational approach to rap lyrics generation. In: Proceedings of the ACM
  SIGKDD International Conference on Knowledge Discovery and Data Mining, pp
  195--204

\bibitem[{Mauch et~al(2015)Mauch, MacCallum, Levy, and
  Leroi}]{mauch2015evolution}
Mauch M, MacCallum RM, Levy M, Leroi AM (2015) The evolution of popular music:
  {USA} 1960--2010. Royal Society open science 2(5)

\bibitem[{Mikolov et~al(2013)Mikolov, Chen, Corrado, and
  Dean}]{mikolov2013efficient}
Mikolov T, Chen K, Corrado G, Dean J (2013) Efficient estimation of word
  representations in vector space. arXiv preprint arXiv:13013781

\bibitem[{Oliveira(2012)}]{oliveira2012poetryme}
Oliveira HG (2012) {PoeTryMe}: a versatile platform for poetry generation. In:
  Proceedings of Workshop on Computational Creativity, Concept Invention, and
  General Intelligence

\bibitem[{Potash et~al(2015)Potash, Romanov, and
  Rumshisky}]{Potash2015GhostWriter}
Potash P, Romanov A, Rumshisky A (2015) Ghostwriter: Using an {LSTM} for
  automatic rap lyric generation. In: Proceedings of the Conference on
  Empirical Methods in Natural Language Processing, pp 165--177

\bibitem[{Potash et~al(2016)Potash, Romanov, and
  Rumshisky}]{potash2016evaluating}
Potash P, Romanov A, Rumshisky A (2016) Evaluating creative language
  generation: The case of rap lyric ghostwriting. arXiv preprint
  arXiv:161203205

\bibitem[{Quoc and Tomas(2014)}]{quoc2014distributed}
Quoc VL, Tomas M (2014) Distributed representations of sentences and documents.
  In: Proceedings of the 31st International Conference on Machine Learning, pp
  1188--1196

\bibitem[{Real and Vargas(1996)}]{Real1996The}
Real R, Vargas JM (1996) The probabilistic basis of {Jaccard's} index of
  similarity. Systematic Biology 45(3):380--385

\bibitem[{Ruli et~al(2012)Ruli, Graeme, and Henry}]{ruli2012using}
Ruli M, Graeme R, Henry sT (2012) Using genetic algorithms to create meaningful
  poetic text. Journal of Experimental \& Theoretical Artificial Intelligence
  24(1):43--64

\bibitem[{Thorsten(2006)}]{thorsten2006training}
Thorsten J (2006) Training linear {SVMs} in linear time. In: Proceedings of the
  12th {ACM} {SIGKDD} International Conference on Knowledge Discovery and Data
  Mining, pp 217--226

\bibitem[{Tsaptsinos(2017)}]{Tsaptsinos2017lyrics}
Tsaptsinos A (2017) Lyrics-based music genre classification using a
  hierarchical attention network. In: Proceedings of the 18th International
  Society for Music Information Retrieval Conference, pp 694--701

\bibitem[{Wang et~al(2016)Wang, Luo, Wang, and Xing}]{wang2016chinese-song}
Wang Q, Luo T, Wang D, Xing C (2016) Chinese song iambics generation with
  neural attention-based model. In: Proceedings of the 25th International Joint
  Conference on Artificial Intelligence, pp 2943--2949

\bibitem[{Wu et~al(2013)Wu, Addanki, Saers, and Beloucif}]{wu2013learning}
Wu D, Addanki VSK, Saers MS, Beloucif M (2013) Learning to freestyle: Hip hop
  challenge-response induction via transduction rule segmentation. In:
  Proceedings of the 2013 Conference on Empirical Methods in Natural Language
  Processing, pp 102--112

\bibitem[{Yan et~al(2013)Yan, Jiang, Lapata, Lin, Lv, and Li}]{rui2013i}
Yan R, Jiang H, Lapata M, Lin SD, Lv X, Li X (2013) i, {Poet}: Automatic
  chinese poetry composition through a generative summarization framework under
  constrained optimization. In: Proceedings of International Joint Conference
  on Artificial Intelligence, pp 2197--2203

\bibitem[{Yu and Mohammed(2017)}]{chen2017kate}
Yu C, Mohammed JZ (2017) {KATE}: K-competitive autoencoder for text. In:
  Proceedings of the ACM SIGKDD International Conference on Data Mining and
  Knowledge Discovery, pp 85--94

\end{thebibliography}

\end{document}